\documentclass[a4paper,fleqn]{cas-dc}
\usepackage[numbers]{natbib}
\usepackage{subcaption}
\usepackage{amsmath}
\usepackage{bm}

\def\tsc#1{\csdef{#1}{\textsc{\lowercase{#1}}\xspace}}
\tsc{WGM}
\tsc{QE}

\begin{document}
\let\WriteBookmarks\relax
\def\floatpagepagefraction{1}
\def\textpagefraction{.001}
\let\printorcid\relax 

\shorttitle{}    

\shortauthors{Sen Peng et al.}

\title[mode = title]{RMAvatar: Photorealistic Human Avatar Reconstruction from Monocular Video Based on Rectified Mesh-embedded Gaussians}

\author[1]{Sen Peng}
\author[2]{Weixing Xie}
\author[3]{Zilong Wang}
\author[3]{Xiaohu Guo}
\author[4]{Zhonggui Chen}
\author[1]{Baorong Yang}
\cormark[1]
\author[5]{Xiao Dong}
\cormark[1]
\address[1]{College of Computer Engineering, Jimei University, Xiamen, China}
\address[2]{National Institute for Data Science in Health and Medicine, Xiamen University, Xiamen, China} 
\address[3]{Department of Computer Science, The University of Texas at Dallas, Richardson, United States}
\address[4]{School of Informatics, Xiamen University, Xiamen, China}
\address[5]{Guangdong Provincial/Zhuhai Key Laboratory of IRADS, BNU-HKBU United International College, Zhuhai, China} 
\cortext[1]{Corresponding authors. Email addresses:\newline yangbaorong@jmu.edu.cn, xiaodong@uic.edu.cn.}  

\begin{abstract}
We introduce RMAvatar, a novel human avatar representation with Gaussian splatting embedded on mesh to learn clothed avatar from a monocular video. We utilize the explicit mesh geometry to represent motion and shape of a virtual human and implicit appearance rendering with Gaussian Splatting. Our method consists of two main modules: Gaussian initialization module and Gaussian rectification module. We embed Gaussians into triangular faces and control their motion through the mesh, which ensures low-frequency motion and surface deformation of the avatar. Due to the limitations of LBS formula, the human skeleton is hard to control complex non-rigid transformations. We then design a pose-related Gaussian rectification module to learn fine-detailed non-rigid deformations, further improving the realism and expressiveness of the avatar. We conduct extensive experiments on public datasets, RMAvatar shows state-of-the-art performance on both rendering quality and quantitative evaluations. Please see our project page at \href{https://rm-avatar.github.io}{https://rm-avatar.github.io}.
\end{abstract}




\maketitle

\section{Introduction}

High-fidelity animatable human avatar modeling from  video has been a longstanding challenge in computer vision and graphics. Rendering photorealistic avatars from arbitrary views holds importance due to its wide applications in telepresence~\cite{kachach2020virtual}, gaming~\cite{zackariasson2012video}, movie making and AR/VR~\cite{weidner2023systematic}. Modeling human avatars from video requires fusing multiple 2D observations to synthesize a 3D consistent human model. Traditional methods usually rely on dense multi-view supervision to reconstruct the avatar, in most actual scenarios, such a complex multi-view camera system~\cite{joo2015panoptic,wuu2022multiface} is not readily available. The under-constrained nature of monocular observation makes the task of reconstructing the unseen poses and viewpoints of human avatars more challenging. In addition, the distortion of clothes, hair and hand movements are also difficult to reconstruct, and the rendering quality of these parts will affect the realism of the avatar.

Recent methods~\cite{geng2023learning,jiang2022neuman,peng2021animatable,weng2022humannerf} based on implicit neural fields~\cite{mildenhall2021nerf,niemeyer2020differentiable,oechsle2021unisurf} usually learn a canonical avatar representation by mapping camera rays from observation space to canonical space. 
In NeRF framework~\cite{pumarola2021d,weng2022humannerf}, the inverse mapping may project multiple points from the observation space onto same point in the canonical space. This ambiguous correspondence affects the rendering quality, especially for objects with significant motion or high-frequency details. In addition, the heavy MLPs used to model the underlying neural radiation fields are computationally expensive, resulting in long training and inference time. Other than implicit neural representations, point-based methods~\cite{wiles2020synsin,ruckert2022adop, zheng2023pointavatar,lassner2021pulsar,yifan2019differentiable,zhang2022differentiable} are efficient and can capture flexible topology, but may produce incomplete surface geometries.

The recent 3D Gaussian Splatting method~\cite{kerbl20233d} surpasses NeRF in both rendering quality and efficiency by optimizing discrete 3D Gaussian primitives to learn explicit representation of the scene. Follow-up works~\cite{chen2024monogaussianavatar,hu2024gaussianavatar} utilize the strong representation ability of 3D Gaussians and reconstruct human avatar via pose-dependent appearance modeling. Although the current work has made substantial progress, the authenticity of avatar needs to be further improved. Existing human model reconstruction works can be roughly divided into two categories. One is implicit neural avatar based on Gaussian Splatting~\cite{hu2024gaussianavatar,qian20243dgs,hu2024gauhuman}. Given initial Gaussian primitives of avatar, these methods first learn 3D Gaussians in the canonical space, and transform Gaussians to the observation space based on the guidance of human pose and LBS~\cite{lewis2023pose} . These methods adopt the multi-layer perceptrons (MLP) for motion control~\cite{qian20243dgs,xu2024gaussian,xiang2024flashavatar}, which is inferior to mesh-based representation to capture surface deformation. The other category is the hybrid avatar representation~\cite{shao2024splattingavatar,qian2024gaussianavatars,wen2024gomavatar}, which combines rendering quality of Gaussian splatting with geometry modeling of deformable meshes. Specifically, a Gaussian is attached to a mesh face and deformed with the face. The hybrid representation enables a compact and topology complete avatar, thus providing better regularization for Gaussians in novel poses. This representation is beneficial for learning avatars under monocular observation. However, the flexibility of the model needs to be improved. In addition to obtaining the avatar in the observation space based on the mesh guidance, the Gaussian primitives need to be further fine-tuned to enhance the ability to learn complex personal characters, such as twisted clothes and wispy hair.

To address these issues, we introduce a novel 3D avatar representation which is designed to model personalized human avatar with complex identity and motion. Our model, namely RMAvatar, is a hybrid 3D representation with Gaussians embedded on a mesh. The use of the mesh can more accurately represent body motion and surface deformation, providing accurate positioning of the Gaussians in observation space. Our method consists of two main modules: Gaussian initialization module and Gaussian rectification module. At the first, we load a template mesh and deform it to obtain posed mesh at a certain frame. Our method binds 3D Gaussian splats to the posed mesh locally and transform the splats to global space. The Gaussian splat now has a good initial position and other default properties. We optimize the Gaussians by minimizing color loss on the rendering. However, methods based on LBS formula
deformation cannot capture the motion of fine non-surface
regions, such as cloth distortions, hair and skin winkles. Thus the representation ability of Gaussians on posed mesh need to be further improved. We then propose the Gaussian rectification strategy, which is a pose-dependent non-rigid deformation module based on MLP. This module allows us to predict further positional adjustments and covariance shifts, significantly boosting the avatar’s realism and expressiveness. 

In summary, the contributions of our method are as follows:

\begin{itemize}
    \item We propose RMAvatar to model personalized high-fidelity human avatar from monocular videos based on mesh-embedded Gaussian splats.
    \item We design Gaussian rectification module to accurately capture complex non-rigid deformation relate to pose to improve the realism of the avatar.
    \item We conduct extensive experiments on public datasets to demonstrate the superior reconstruction ability of our method on both rendering quality and quantitative evaluations.
\end{itemize}

\begin{figure*}
\centering
\includegraphics[width=1\textwidth]{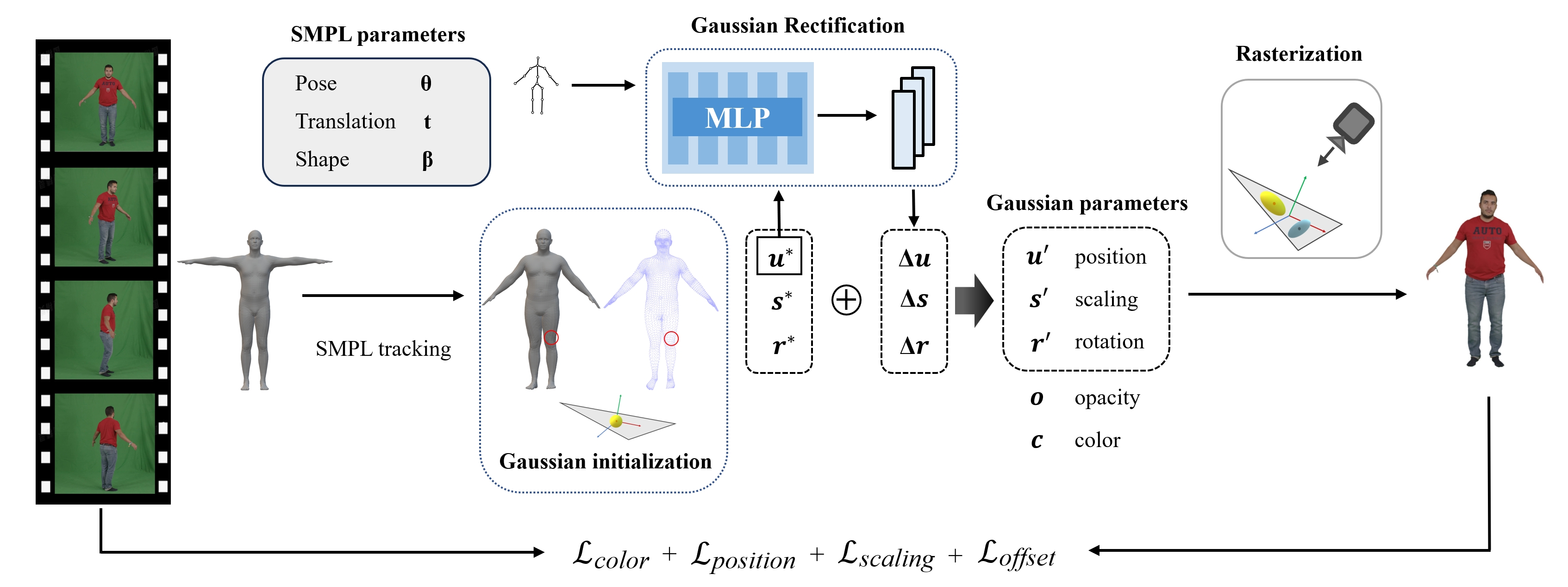} 
\caption{Overview of RMAvatar. Given a sequence of monocular images and a neutral SMPL template, we first obtain the deformed mesh under current pose of the person via SMPL tracking. Our initialization strategy is to embed Gaussian splats on mesh in the local coordinates of each triangle and then transform them to world space based on triangle's shape. To improve the representation ability of Gaussian splats for non-rigid cloth deformation, the rectification module is designed to further adjust Gaussians to learn pose-dependent appearance details of avatar. Finally, the Gaussians in observation frame and their respective color values are accumulated via differentiable Gaussian rasterization to render the image.}
\label{fig1:pipline}
\end{figure*}

\section{Related work}

Animatable human avatar reconstruction from monocular videos is challenging in capturing high-quality geometry deformation and appearance. Early works take parametric template models, e.g., FLAME~\cite{FLAME:SiggraphAsia2017} and SMPL family~\cite{SMPL:2015,SMPL-X:2019}, which provide vertices with fixed connectivity as an explicit prior of the 3D human avatar. By unwrapping them to a unified UV space, texture atlas can be obtained through differentiable rendering~\cite{niemeyer2020differentiable}. These explicit mesh-based models can be easily fit into existing rendering pipeline and the vertices of mesh templates can be easily deformed to capture pose-dependent geometry deformation. However, since the mesh templates don't model clothes and have fixed topology, these methods often suffer from capturing fine-scale deformation of the human body, especially the distortion and color changes of clothes or movements of hairs or local details in the face. To address these problems, researchers have investigated ways, e.g., via learning 3D vertex offsets~\cite{alldieck2019learning,moreau2024human,wang2022arah,PhysAavatar24} for clothes, or resorting to implicit representations, e.g., neural radiance fields (NeRF) ~\cite{jiang2022neuman,weng2022humannerf,icsik2023humanrf} and Gaussian fields~\cite{hu2024gaussianavatar, chen2024monogaussianavatar,qian20243dgs,hu2024gauhuman,chen2023monogaussianavatar,lin2024layga}. 
\subsection{Implicit Function-based human avatar}
Implicit models normally encode the 3D human avatar with implicit surface functions~\cite{peng2021neural}, e.g., SDF~\cite{jiang2022selfrecon,zheng2022avatar,xu2024relightable}, occupancy field~\cite{mihajlovic2021leap, chen2023fast,li2022avatarcap} and NeRF~\cite{icsik2023humanrf, weng2022humannerf, jiang2022neuman, peng2021animatable, jiang2023instantavatar, yu2023monohuman,chen2021animatable}. NeRF-based methods learns the neural radiance fields of human from videos and render novel views with differentiable volumetric rendering. These methods present a deformable NeRF representation by unwrapping different poses to a shared canonical space with inverse kinematic transformations as well as residual deformations for modelling animatable human avatar from videos. Especially, HumanNeRF~\cite{weng2022humannerf} models human motion by decomposing it into skeletal and non-rigid deformations and refines texture details by aggregating color and depth information from neighboring views, showing its impressive results on view-synthesis for human avatar. Anim-NeRF~\cite{chen2021animatable} learns pose conditioned inverse LBS field to capture the fine details of human. However, the implicit function-based methods usually adopt pure MLPs to model the human avatar, yielding smooth and blurry quality and low rendering speed~\cite{peng2021animatable}. To address the expensive computation of NeRF-based animatable human avatar, efforts has been made in accelerated data structures for speeding up training and inference of NeRFs~\cite{wang2022fourier, zhao2022human, jiang2023instantavatar, geng2023learning, kwon2024deliffas}. However, some works rely on dense multi-view inputs~\cite{wang2022fourier, zhao2022human, kwon2024deliffas,li2024animatable,chen2021mvsnerf,chibane2021stereo,xu2024gphm} to achieve good rendering quality of animatable human avatar. Instant-NVR~\cite{geng2023learning} use iNGP~\cite{muller2022instant} as the underlying representation for articulated NeRFs, and modelling non-rigid deformations in the UV space for fast training. However, Instant-NVR generates blurring renderings on the non-rigid deformations~\cite{qian20243dgs}.

\subsection{Point-based Human Avatar}
Point cloud is also a commonly used representation for human avatar. DPF~\cite{prokudin2023dynamic} and NPC~\cite{su2023npc} apply Point-NeRF~\cite{xu2022point} for producing explicit surface points and learning non-rigid deformation of human avatars from RGB videos. However, with the MLPs used in Point-NeRF, these methods still struggle with the blurry rendering results and the computation efficiency problem. Point-based rendering~\cite{zheng2022structured, zheng2023pointavatar, kerbl20233d} has been adopted as an efficient alternative to NeRFs. PointAvatar~\cite{zheng2023pointavatar} takes the advantage of using explicit point primitives in forward rasterization and learns a forward deformation from canonical to pose space, generating high-fidelity animatable 3D avatars from monocular videos. As an explicit point-based 3D representation, 3D Gaussian Splatting (3DGS)~\cite{kerbl20233d} has shown its efficiency in real-time photo-realistic rendering of static scenes. Concurrent works~\cite{hu2024gauhuman,kocabas2024hugs,lei2024gart,li2024animatable,qian2024gaussianavatars,qian20243dgs,shao2024control4d,wang2024gaussianeditor,zheng2024gps} also introduce 3DGS into animatable human avatars. 
GaussianAvatar~\cite{hu2024gaussianavatar} introduces animatable 3D Gaussians to explicitly represent human in various poses and clothing styles, and jointly optimizes motions and appearances during avatar modeling to tackle the issue of inaccurate motion estimation from monocular video. 3DGS-Avatar~\cite{qian20243dgs} leverages 3DGS~\cite{kerbl20233d} and efficiently learns a non-rigid transformation network to reconstruct animatable clothed human avatars. GART~\cite{lei2024gart} utilizes a mixture of moving 3D Gaussians to explicitly approximate a deformable subject's geometry and appearance.

\subsection{Hybrid Human Avatar}
Hybrid 3D representations have also been used in modelling human avatars from videos~\cite{habermann2023hdhumans,chen2024meshavatar}. HDHumans~\cite{habermann2023hdhumans} learns the deformation embedded as NeRF by the parameterized pose and embedded graph of template mesh prior. DELTA~\cite{feng2023learning} propose to hybridly model human avatar with textured mesh for body and NeRF for hair and clothing. SplattingAvatar~\cite{shao2024splattingavatar} proposes a hybrid avatar representation of 3D Gaussians and mesh to disentangle the human motion and appearance. Specifically, the pose-dependent deformation are explicitly defined by mesh, and the geometry and appearance details are modeled by the Gaussians. Similarly, GoMAvatar~\cite{wen2024gomavatar} introduces the Gaussian-on-Mesh representation that leverages 3D Gaussians for real-time rendering. GaussianAvatars~\cite{qian2024gaussianavatars} reconstructs head avatars by rigging 3D Gaussians to a parametric morphable face model with the binding inheritance strategy in which the Gaussian is parameterized with the index of its parent triangle. However, in SplattingAvatar~\cite{shao2024splattingavatar} and GaussianAvatars~\cite{qian2024gaussianavatars}, the Gaussian attributes are independent of the specific pose or expression of the avatar, so non-rigid deformations related to pose cannot be modeled. To solve this problem, based on mesh-embedded Gaussians, we design a Gaussian rectification module to help Gaussians represent areas that SMPL~\cite{SMPL:2015} cannot model, such as clothes.


\section{Method}

Figure~\ref{fig1:pipline} shows the pipeline of our method RMAvatar. Given a sequence of monocular images and a SMPL template, we deform the mesh to a certain pose of a person via LBS. Each Gaussian is embedded on one triangle of the mesh. The 3D Gaussian is relatively static to its parent triangle but dynamic in the global space as the triangle moves. The hybrid representation of human avatar via Gaussians on mesh defines the position of the Gaussians in posed space.  Except for position, each gaussian has its attributes of rotation, scaling, opacity and color. To reconstruct personalized high-fidelity avatars, we add pose-dependent transformations to each Gaussian to learn details caused by non-rigid transformations of the human avatar.



\subsection{Preliminary}

3D Gaussian Splatting~\cite{kerbl20233d} employs a set of anisotropic Gaussian primitives to explicitly represent a static 3D scenes.
Each Gaussian splat is characterized by a covariance matrix $\Sigma$ at position $\mathbf{\mu}$, which is referred as the mean of Gaussian:
\begin{equation}
\label{formula:gaussian}
    G(x)=e^{-\frac{1}{2}(x-\mu)^T\Sigma^{-1}(x-\mu)}.
\end{equation}
To ensure positive semi-definite nature of the covariance matrix, it can be decomposed into a scaling matrix $S$ and a rotation matrix $R$:
\begin{equation}
\label{formula:covariance decomposition}
    \Sigma = RSS^TR^T.
\end{equation}
In practice, we store diagonal vector $s \in \mathbb{R}^3$ of scaling matrix $S$ and a quaternion vector $r \in \mathbb{R}^4$ of rotation matrix $R$ for a Gaussian. 

During the rendering, the 3D Gaussians are projected to 2D image plane and accumulated via alpha blending.
As introduced by~\cite{zwicker2001ewa}, using a viewing transform matrix $W$ and the Jacobian matrix $J$ of the affine approximation of the projective transformation, the covariance matrix $\Sigma^{\prime}$ in 2D camera space can be computed as:
\begin{equation}
    \Sigma^{\prime} = JW\Sigma W^TJ^T.
\end{equation}
The color of a pixel is then calculated by blending 3D Gaussian splats that overlap at that pixel, and these Gaussians are sorted according to their depth:
\begin{equation}
\label{formula: splatting color}
    C = \sum_{i\in N}c_i \alpha_i \prod_{j=1}^{i-1} (1-\alpha_j).
\end{equation}
Here, $\alpha_i$ is blending weight calculated by opacity $\sigma_i$ multiplied with the probability density of projected 2D Gaussian at the target pixel location. $c_i$ is the view-dependent color of Gaussian $G_i$ represented by spherical harmonic coefficients. 

We denote Gaussian properties as $G = \left\{\mu, r, s, \sigma, c \right\}$. After rasterization, the Gaussian properties are optimized through appearance and other losses to obtain a 3D representation of the scene. In addition, the adaptive control of the Gaussian can improve its representation ability, mainly including three operations: splitting, cloning and pruning. Splitting and cloning are performed on Gaussians with large position gradients, which augments the number of Gaussians. Pruning operation periodically eliminates Gaussians with excessively small opacity to suppress floating artifacts.

\subsection{Gaussian Initialization on Mesh}
Our method is a hybrid representation of avatar with Gaussians embedded on mesh~\cite{qian2024gaussianavatars,shao2024splattingavatar,wen2024gomavatar}. Given a set of monocular images, the registered mesh corresponding to each frame can be obtained by deforming the neutral SMPL template via shape and pose parameters. The mesh consists of a set of vertices $V = \left\{v_i\right\}_1^n$ and faces $F = \left\{f_j\right\}_1^m$.

Given a triangle with vertices $(v_a, v_b, v_c)$ and normal $\bm{n}$, we bind a Gaussian to the triangle. Specifically, we construct a local coordinate system with the mean position $\bm{M}$ of three vertices as the origin. The local coordinate system is determined by an edge vector $\bm{e} = v_b-v_a$, the normal of the triangle $\bm{n}$, and their cross product $\bm{e}\times \bm{n}$. These three vectors constitute a rotation matrix $\bm{R}$ of the triangle in the global space, describing the orientation of the triangle. The scaling $\bm{w}$ of triangle is measured by the mean length of one edge $\bm{e}$ and its perpendicular $\bm{e_p}$. 

We initialize the position of the Gaussian on triangle to the local origin, the rotation $\bm{r}$ to identity matrix, and the scaling to unit vector. We then convert these properties from local coordinate system of the triangle to the global coordinate system. The global position, rotation and scaling of the Gaussian $\mu^*$, $\bm{r}^*$ and $s^*$ are computed by:
\begin{align}
    \mu^* & = \bm{w}\bm{R}\mu + \bm{M},\\
    \bm{r}^* & = \bm{R}\bm{r} \label{eq:rotation},\\
     s^* & = \bm{w}s. 
\end{align}

Assigning a single Gaussian to each triangle is not enough to capture complex details. For example, there may be only one triangle under the distorted clothing or curly hair, and a single Gaussian splat on it is not enough to represent such complex appearance. We adaptively adjust the density of Gaussian splats on mesh using strategies such as splitting, cloning and pruning. Specifically, the splitting operation is performed on Gaussians with magnitude of scaling matrix larger than a threshold, and cloning operation is performed with magnitude of scaling matrix smaller than a threshold. The pruning operation is periodically applied to reset the opacity of all Gaussians close to zero and removes Gaussians with opacity below a threshold. A Gaussian stores the index of its parent triangle, and we make sure that every triangle has at least one Gaussian attached to it.

\subsection{Gaussian Rectification for Non-rigid Deformation}
Gaussian motion guided by triangle mesh can capture rough human movements, but constrained by the linear representation of LBS, Gaussians that move with the mesh are insufficient for capturing non-rigid distortions and intricate dynamic textures. We decompose complex human motion to rigid transformation guided by human skeleton and non-rigid transformation caused by pose-dependent cloth distortions~\cite{weng2022humannerf}. It is necessary to design a separate module for non-rigid deformation to further adjust the Gaussian properties dependent on poses.

Given a human pose $\bm{P}$ in current observation and Gaussian positions $\mu^*$ on the posed mesh, we formulate the non-rigid deformation module $\mathcal{F}_\theta$ based on MLP as: 

\begin{equation}
\label{delta}
    (\delta \mu, \delta r, \delta s) = \mathcal{F}_\theta(\gamma(\mu^*),\bm{P}).
\end{equation}

We use an embedding function $\gamma(\cdot)$ to encode Gaussian positions with a specific frequency. The change of Gaussian attributes is related to the pose of each frame, which enables Gaussian to learn the non-rigid changes caused by pose, making up for the shortcomings of LBS direct linear representation and further improving the realism of avatar. Based on the predicted offset, the rectified Gaussian attributes are calculated as follows:
\begin{align}
\mu^\prime & = \mu^* + \delta \mu,\\
r^\prime & = r^* \cdot \delta r,\\
s^\prime & = s^* + \delta s.
\end{align}

Here, the updated $\mu^\prime$, $r^\prime$ and $s^\prime$ are in global space. $r^*$ is the quaternion vector corresponding to the rotation in Equation~\ref{eq:rotation}. Applying $\cdot$ operation to the quaternion vectors is equivalent to multiplying the corresponding rotation matrices. Combined with opacity and color, we get the Gaussian property set $G = \left\{\mu^\prime, r^\prime, s^\prime, \sigma, c \right\}$ for subsequent rasterization.

Based on above analysis, the motion guidance of the mesh ensures that Gaussians learn the geometry of the avatar, while the pose-based recitification enables the Gaussians to model the appearance changes dependent on pose. By modeling both rigid and non-rigid deformations, our approach is able to reconstruct more realistic avatars.


\begin{table*}[t]
\centering
\caption{ Quantitative comparison on PeopleSnapshot~\cite{alldieck2018video}. Compared with state-of-the-art methods, our method achieves significant improvement in rendering quality of novel view synthesis on all evaluation metrics. The best results are in bold.}
\resizebox{\textwidth}{!}{
\begin{tabular}{lcccccccccccc}
    \toprule
    & \multicolumn{3}{c}{male-3-casual} & \multicolumn{3}{c}{male-4-casual} & \multicolumn{3}{c}{female-3-casual} & \multicolumn{3}{c}{female-4-casual}  \\
    & PSNR$\uparrow$ & SSIM$\uparrow$ & LPIPS$\downarrow$ & PSNR$\uparrow$ & SSIM$\uparrow$ & LPIPS$\downarrow$ & PSNR$\uparrow$ & SSIM$\uparrow$ & LPIPS$\downarrow$ & PSNR$\uparrow$ & SSIM$\uparrow$ & LPIPS$\downarrow$ \\
    \midrule
    Anim-NeRF~\cite{chen2021animatable} & 29.37 & 0.970 & 0.017 & 28.37 & 0.960 & 0.027 & 28.91 & 0.974 & 0.022 & 28.90 & 0.968 & 0.017 \\
    InstantAvatar~\cite{jiang2023instantavatar} & 30.91 & 0.977 & 0.022 & 29.77 & 0.980 & 0.025 & 29.73 & 0.975 & 0.025 & 30.92 & 0.977 & 0.021 \\
    GaussianAvatar~\cite{hu2024gaussianavatar} & 30.98 & 0.979 & 0.015 & 28.78 & 0.975 & 0.023 & 29.55 & 0.976 & 0.023 & 30.84 & 0.977 & 0.014 \\
    SplattingAvatar~\cite{shao2024splattingavatar} & 32.31 & 0.978 & 0.031 & 30.51 & 0.978 & 0.041 & 30.42 & 0.976 & 0.044 & 31.12 & 0.976 & 0.032 \\
    \midrule
    Ours & \textbf{34.12} & \textbf{0.985} & \textbf{0.013} & \textbf{31.23} & \textbf{0.983} & \textbf{0.022} & \textbf{31.42} & \textbf{0.980} & \textbf{0.021} & \textbf{33.06} & \textbf{0.982} & \textbf{0.013} \\
    \bottomrule
\end{tabular}
}
\label{tab:exp_people}
\end{table*}

\begin{table*}[t]
\centering
\caption{Quantitative comparison on ZJU-MoCap~\cite{peng2021neural}. Our method achieves best reconstruction quality on novel view synthesis of avatars with large motions. The best results are in bold.
}
\resizebox{\textwidth}{!}{
\begin{tabular}{lcccccccccccccccccc}
    \toprule
    & \multicolumn{3}{c}{377} & \multicolumn{3}{c}{386} & \multicolumn{3}{c}{387} & \multicolumn{3}{c}{392} & \multicolumn{3}{c}{393} & \multicolumn{3}{c}{394}  \\
    & PSNR$\uparrow$ & SSIM$\uparrow$ & LPIPS$\downarrow$& PSNR$\uparrow$ & SSIM$\uparrow$ & LPIPS$\downarrow$& PSNR$\uparrow$ & SSIM$\uparrow$ & LPIPS$\downarrow$& PSNR$\uparrow$ & SSIM$\uparrow$ & LPIPS$\downarrow$ & PSNR$\uparrow$ & SSIM$\uparrow$ & LPIPS$\downarrow$ & PSNR$\uparrow$ & SSIM$\uparrow$ & LPIPS$\downarrow$\\
    \midrule
    GaussianAvatar~\cite{hu2024gaussianavatar} & 24.86 & 0.944 & 0.063 & 27.10 & 0.923 & 0.074 & 25.63 & 0.947 & 0.043 & 26.18 & 0.929 & 0.088 & 23.90 & 0.919 & 0.099 & 26.11 & 0.925 & 0.084 \\
    SplattingAvatar~\cite{shao2024splattingavatar} & 32.24 & 0.977 & 0.028 & 30.31 & 0.953 & 0.073 & 30.69  & 0.967  & 0.045  & 33.41 & 0.975 & 0.040  & 30.02  & 0.962 & 0.047 & 32.36 & 0.965 & 0.044  \\
    \midrule
    Ours & \textbf{32.68} & \textbf{0.982} & \textbf{0.015}  & \textbf{30.61} & \textbf{0.955} & \textbf{0.055} & \textbf{31.01} & \textbf{0.973} & \textbf{0.021} & \textbf{34.30} & \textbf{0.979} & \textbf{0.024} & \textbf{31.09} & \textbf{0.969} & \textbf{0.025} & \textbf{32.70} &  \textbf{0.969} &   \textbf{0.025} \\
    \bottomrule
\end{tabular}
}
\label{tab:exp_zju}
\end{table*}

\subsection{Optimization}

We utilize RGB loss $\mathcal{L}_{rgb}$, SSIM loss~\cite{wang2004image} $\mathcal{L}_{ssim}$ and LPIPS loss~\cite{zhang2018unreasonable} $\mathcal{L}_{lpips}$ with the corresponding weights $\lambda_{rgb}$, $\lambda_{ssim}$ and $\lambda_{lpips}$ to optimize the rendered images: 
\begin{equation}
\label{loss_rgb}
    \mathcal{L}_{color} = \lambda_{rgb}\mathcal{L}_{rgb}+\lambda_{ssim}\mathcal{L}_{ssim} +\lambda_{lpips}\mathcal{L}_{lpips}.
\end{equation}
Note that $\mathcal{L}_{rgb}$, $\mathcal{L}_{ssim}$ and $\mathcal{L}_{lpips}$ are $L_1$-norm losses. These losses for rendering quality already result in good reconstruction images. Besides, we apply some regularization terms to constrain the position and covariance learning for Gaussians.

At initialization, we bind a Gaussian to the local origin of the triangle. During training, the position of Gaussian splat is optimized and may deviate away from its parent triangle if not constrained. Large drifts do not affect rendering quality from visible perspectives, but may cause artifacts when animating the Gaussians to a new pose via SMPL. Following~\cite{qian2024gaussianavatars}, we regularize the local position of each Gaussian by:
\begin{equation}
\label{loss_pos}
 \mathcal{L}_{pos} = \| \max \left( \mu- \epsilon_{pos}, 0 \right) \|_2,
\end{equation}
where $\epsilon_{pos}$ is set to $1$ to constrain the Gaussians around their parent triangles and give them certain freedom to adjust position. In addition to the position, the scaling of the Gaussian also affects the rendering quality. A Gaussian that is too large compared to the parent triangle is sensitive to the motion of triangle, thus can easily introduce jitter and artifacts when rotating with the mesh. We regularize the scaling of each Gaussian and utilize the threshhold $\epsilon_{scaling}$ to prevent it from shrinking excessively:
\begin{equation}
\label{loss_scaling}
 \mathcal{L}_{scaling} = \| \max \left( s- \epsilon_{scaling}, 0 \right) \|_2,
\end{equation}
where $\epsilon_{scaling}$ is set to $0.6$ for the maximum allowable scaling. When $\mu$ and $s$ are below
these thresholds, the corresponding loss terms are disabled.

The Gaussians on the mesh are able to capture the rough motion and appearance of the avatar, and the rectification module is proposed only to slightly adjust the Gaussian properties to support its learning of complex details caused by non-rigid deformations. Thus we design a regularization term $\mathcal{L}_{offset}$ to constrain the values of $\delta \mu$, $\delta r$ and $\delta s$:
\begin{equation}
\label{loss_offset}
 \mathcal{L}_{offset} = \| (\delta \mu, \delta r,  \delta s)\|_2.
\end{equation}

\noindent Taking the color losses and regularization losses together, we define the final loss function as follows:
\begin{equation}\label{total loss}
\begin{split}
\mathcal{L} & =  \lambda_{rgb}\mathcal{L}_{rgb}+\lambda_{ssim}\mathcal{L}_{ssim} +\lambda_{lpips}\mathcal{L}_{lpips}\\
& + \lambda_{pos}\mathcal{L}_{pos}+\lambda_{scaling}\mathcal{L}_{scaling} +\lambda_{offset}\mathcal{L}_{offset}.
\end{split}
\end{equation}
Here, the $\mathcal{L}_{pos}$, $\mathcal{L}_{scaling}$ and $\mathcal{L}_{offset}$ are $L_2$-norm losses with corresponding weights $\lambda_{pos}$, $\lambda_{scaling}$ and $\lambda_{offset}$.

\begin{table}[ht]
\centering
\caption{Train/test split of the ZJU-MoCap~\cite{peng2021neural} dataset.}
\begin{tabular}{lcc|lcc}
    \toprule
        & train & test &  & train & test  \\
    \midrule
    377 & 1-456 & 457-617 & 386 & 1-456 & 457-646\\
    387 & 1-456 & 457-654 & 392 & 1-456 & 457-556\\
    393 & 1-456 & 457-658 & 394 & 1-656 & 657-859\\
    \bottomrule
\end{tabular}
\label{tab:zju_split}
\end{table}

\section{Experiments}
In this section, we evaluate RMAvatar on the PeopleSnapshot dataset~\cite{alldieck2018video} and the ZJU-MoCap dataset~\cite{peng2021neural} by comparing with state-of-the-art human avatar modeling methods from monocular videos and systematically ablate each component of RMAvatar. 

\subsection{Datasets and metrics}
\textbf{PeopleSnapshot~\cite{alldieck2018video} Dataset.} We select 4 sequences of the PeopleSnapshot dataset as in InstantAvatar~\cite{jiang2023instantavatar} and follow the same data split. We compare our approach with Anim-NeRF~\cite{chen2021animatable}, InstantAvatar~\cite{jiang2023instantavatar}, GaussianAvatar~\cite{hu2024gaussianavatar} and SplattingAvatar~\cite{shao2024splattingavatar} on this dataset. For fair comparison, we use the provided poses optimized by Anim-NeRF~\cite{chen2021animatable} in our training and inference. 

\begin{figure*}[!t]
    \centering
    \includegraphics[width=0.85\linewidth]{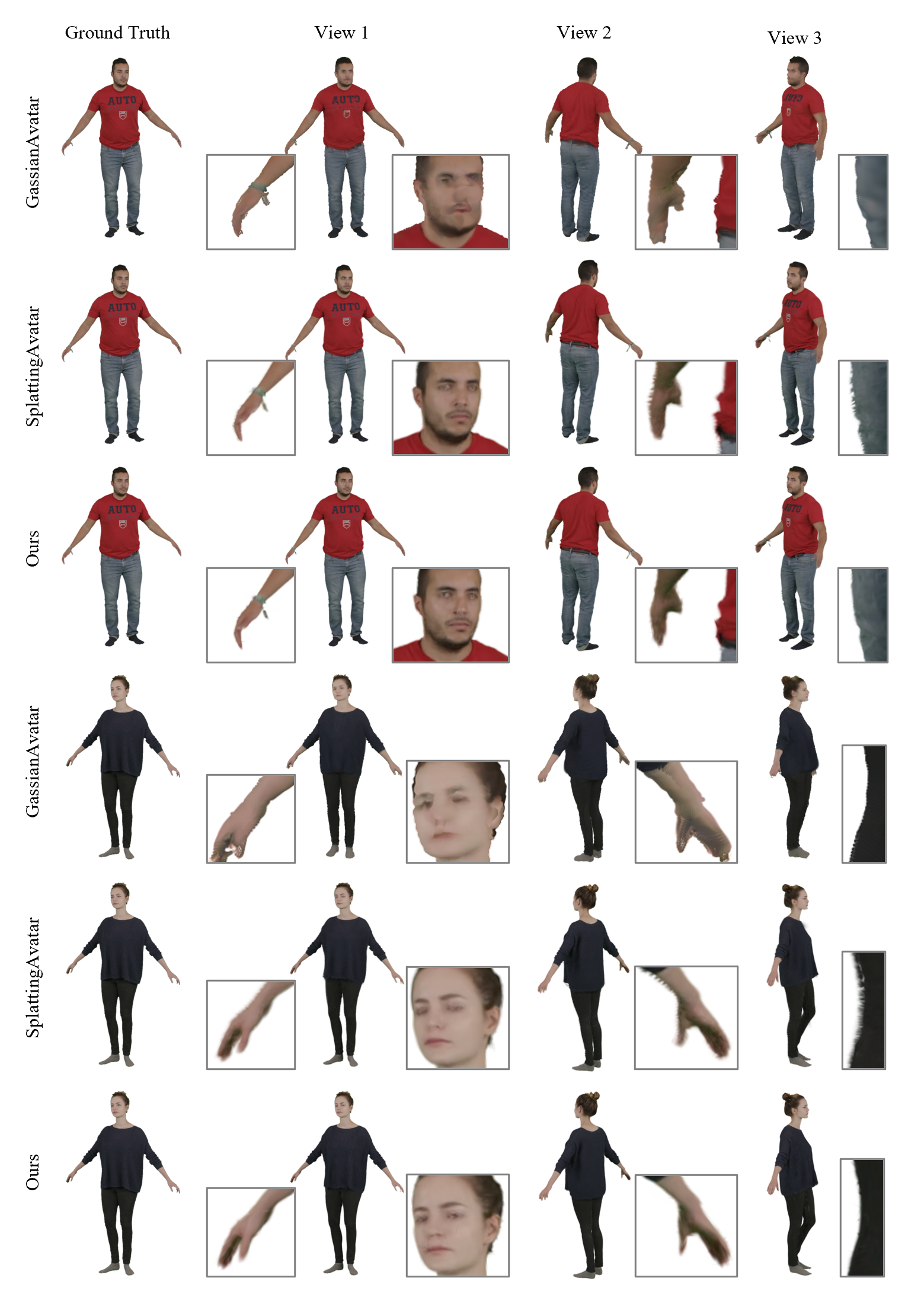}
    \caption{Comparison of novel view synthesis on PeopleSnapshot~\cite{alldieck2018video}. Our method is able to reconstruct intricate texture details.}
    \label{fig:comparison_people}
\end{figure*}

\begin{figure*}[h]
    \centering
    \includegraphics[width=0.85\linewidth]{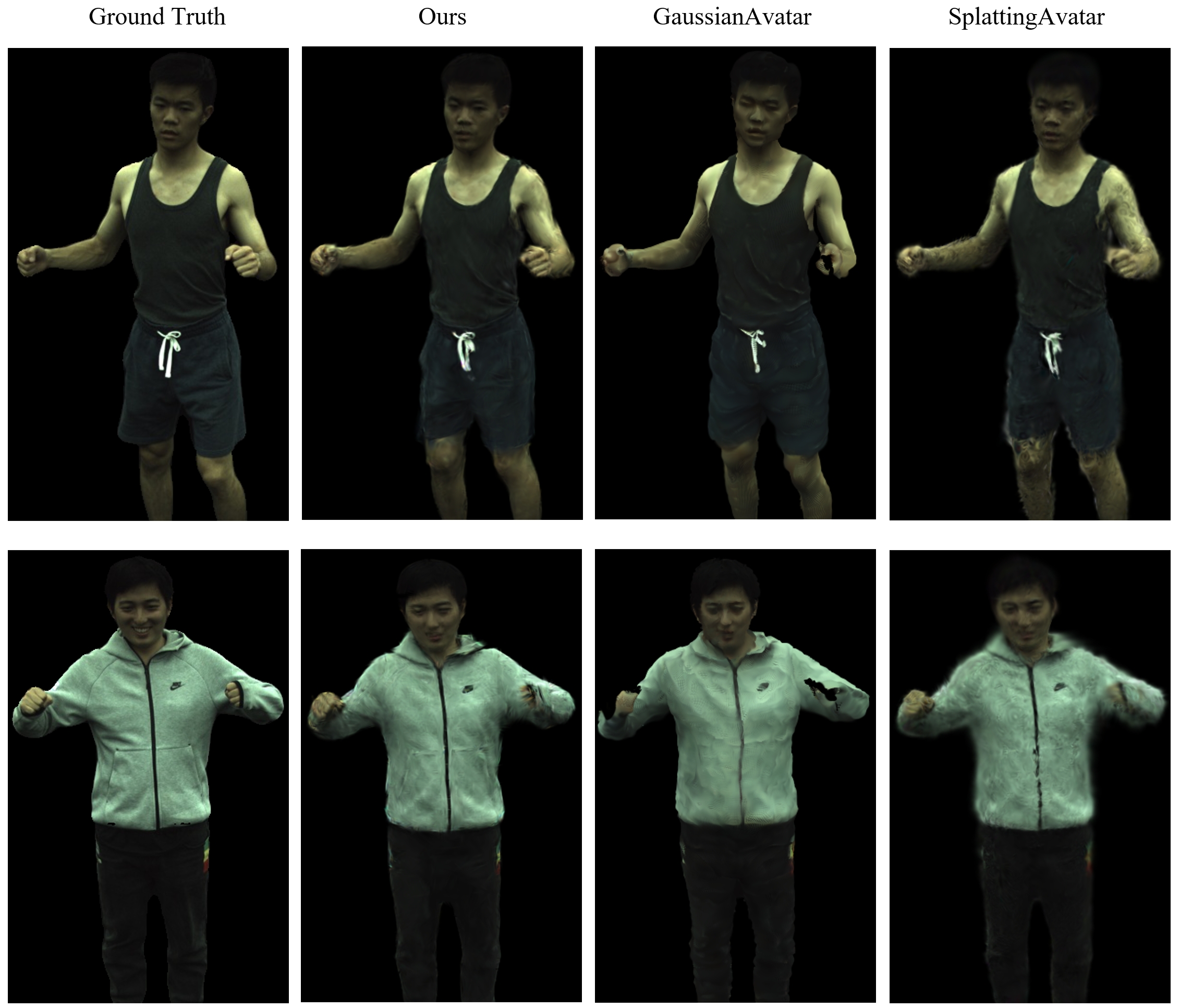}
    \caption{Comparison of novel view synthesis on ZJU-MoCap~\cite{peng2021neural}. Our method reconstructs complicated cloth textures.}
    \label{fig:comparison_zju}
\end{figure*}

\textbf{ZJU-MoCap~\cite{peng2021neural} Dataset.} We use six subjects (377, 386, 387, 392, 393, and 394)~\cite{weng2022humannerf} of the ZJU-MoCap dataset and compare with GaussianAvatar~\cite{hu2024gaussianavatar} and SplattingAvatar~\cite{shao2024splattingavatar} on monocular videos of the dataset. We select a video from a camera viewpoint and use a segment of the video during training. Specifically, we select a clip containing a complete turning action of the avatar as training data. We show the training/test split in Table~\ref{tab:zju_split} and use the same split in comparison with other methods. Note that due to the small motion of the characters, we select 1 frame out of every 4 frames for training. We use Anim-NeRF~\cite{peng2021animatable} to get the refined poses of ZJU-MoCap in our training and inference.

\textbf{Evaluation Metrics.} We consider three metrics: PSNR, SSIM and LPIPS to evaluate the reconstruction quality of different methods on two datasets, which measure the pixel intensity similarity, structural similarity, and perceptual image patch similarity between the rendered and ground truth images.

\begin{figure*}[h]
    \centering
    \includegraphics[width=0.98\linewidth]{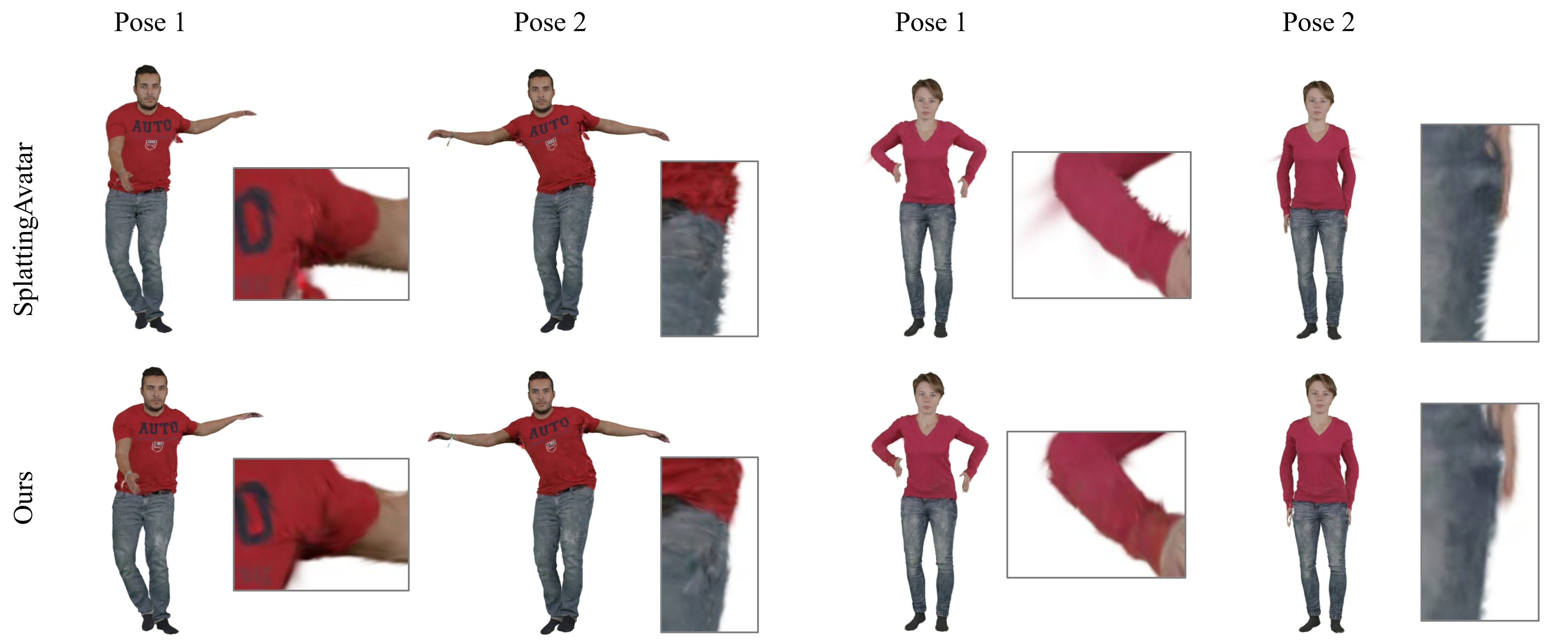}
    \caption{Comparison of animation on out-of-distribution poses on PeopleSnapshot~\cite{alldieck2018video}. Our method generates consistent representations for avatars on novel poses.}
    \label{fig:comparison_novelpose}
\end{figure*}

\subsection{Implementation details}

Our model is trained for 50,000 iterations on a single NVIDIA RTX 3090 GPU, with training time ranging from approximately 50 minutes to an hour. During training, we use Adam~\cite{kinga2015method} to optimize our model. We set the learning rate to $0.008$ for the position and exponentially decay it with a factor of $0.01$ to $10^{-5}$. The learning rate for the scaling, rotation and opacity of 3D Gaussians are 0.017, 0.001 and 0.05 respectively. The learning rate for the Gaussian rectification module is  $10^{-4}$. We use the same configurations as SplattingAvatar~\cite{shao2024splattingavatar} for densification and pruning for the 3D Gaussians. We start to apply density control operations on the Gaussians every 500 iterations, and reset the opacity of Gaussians every 5,000 iterations from iteration 10,000, and turn off the density controller as well as the opacity-rest operation at iteration 35,000. For the target loss, we set the parameters of different loss terms to $\lambda_{rgb} = 1.5$, $\lambda_{ssim} = 0.2$, $\lambda_{lpips}=0.05$, $\lambda_{pos}=0.01$, $\lambda_{scaling}=1$ and $\lambda_{offset}=1$, respectively.

For the Gaussian rectification module, we designed a 5-layer MLP to predict the offsets of Gaussian attributes such as position, rotation, and scaling. The MLP network takes the encoded Gaussian position and the pose parameters as input, which contains 105 channel features. The input channels for subsequent hidden layer are (128, 164, 128), and contains a skip connection in the fourth layer. The output of the last layer of the MLP is a 10-dimensional feature, which contains the position offset $\delta u$, scale offset $\delta s$, and quaternion offset $\delta r$. In order to limit the offset value to be small, we design offset loss for these three attributes.



\subsection{Comparisons on human avatar reconstruction}

We evaluate the reconstruction quality of human avatars by novel view synthesis and avatar animation experiments on two datasets. In Table~\ref{tab:exp_people} and Table~\ref{tab:exp_zju} we report quantitative results of different methods on novel view synthesis on PeopleSnapshot and ZJU-MoCap, respectively. Our method far exceeds the SoTA methods in terms of PSNR, SSIM, and LPIPS, which reflect rendering quality of avatars, indicating that our method can learn the complex textures more clearly. 

\textbf{Novel view synthesis.} We show the novel view synthesis results of GaussianAvatar, SplattingAvatar and our method in Figure~\ref{fig:comparison_people}. From the zoomed-in details of the face and hands, it is obvious that our method produces clear details and smooth boundaries. The contour of the reconstructed human avatar of GaussianAvatar~\cite{hu2024gaussianavatar} is obviously larger than the ground truth, which indicates that the position of the Gaussians it learned are not accurate, and it is difficult to capture big motion and intricate textures, such as human face and cloth wrinkles. The rendering images of SplattingAvatar~\cite{shao2024splattingavatar} are better than GaussianAvatar, which shows the advantages of hybrid avatar representation with Gaussians on mesh. SplattingAvatar embeds Gaussians on canonical mesh and guides the motion of Gaussians by mesh warping from canonical space to posed space. As pointed out by SplattingAvatar, the quality of its reconstruction is highly dependent on the motion accuracy of the underlying mesh. Due to the lack of cloth and hair meshes, they failed to learn clear texture of head and cloth. In addition, this method produces obvious artifacts on the avatar surface due to inaccurate Gaussian positions, which is probably because Walking on mesh strategy does not constrain the amplitude of Gaussian motion effectively. In contrast, our approach designs a separate module to learn non-rigid cloth and hair deformations, which makes up for the defect that Gaussians on SMPL cannot represent clothes, thereby improving the reconstruction accuracy of rendered avatars.

The comparison of  novel view synthesis on the ZJU-MoCap dataset can better reflect the advantages of our method. The avatars have large movements and there are certain errors in estimated poses, which leads to poor visual effects of hybrid methods combining mesh and Gaussian, such as SplattingAvatar. Our method further adjusts the Gaussian properties on the basis of mesh guidance, reduces the reconstruction error caused by inaccurate pose, and ensures the high quality of Avatar.

\textbf{Avatar animation.} In Figure~\ref{fig:comparison_novelpose}, we show the avatar animation results for out-of-distribution poses generated by SplattingAvatar and our method. It is clear that our method demonstrates a more consistent 3D appearance and shape of avatars under challenging novel poses, while SplattingAvatar generates obvious wrong positioned Gaussians and non-smooth boundaries for large motion sequence.

\begin{figure*}[h]
    \centering
    \includegraphics[width=0.98\linewidth]{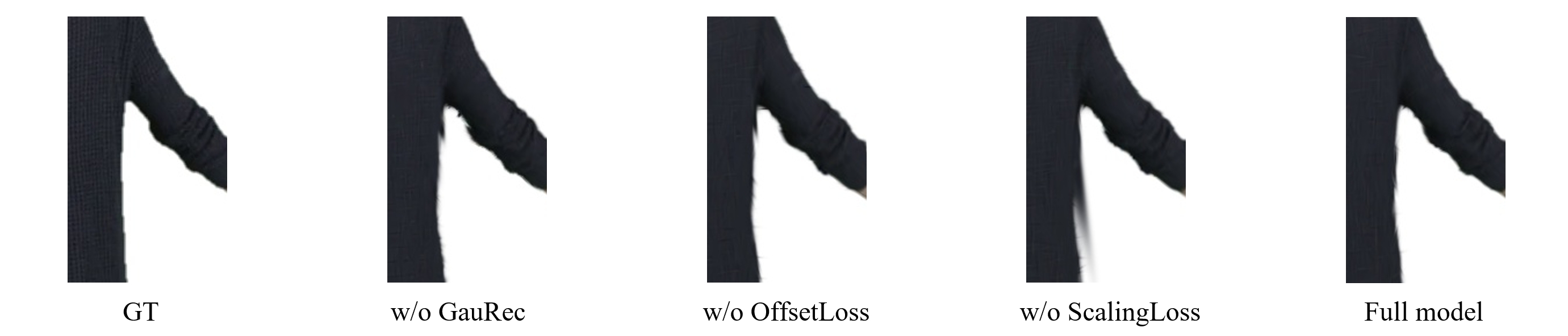}
    \caption{Ablation on female-3-casual of PeopleSnapshot. Our full model mitigates rendering artifacts with Gaussian rectification module and regularization losses.}
    \label{fig:ablation}
\end{figure*}

\subsection{Ablation study}
We conduct ablation studies by comparing the full model with 1) removing the Gaussian rectification module, denoted as ``w/o GauRec'', 2) only removing the regularization term $\mathcal{L}_{offset}$, denoted as ``w/o OffsetLoss'' and 3) only removing the scaling loss term $\mathcal{L}_{scaling}$, denoted as ``w/o ScalingLoss''. The ``w/o GauRec'' model fixes the position of the Gaussian to its parent triangle, and properties such as rotation and scale of the Gaussian remain consistent in all poses. Specifically, the rectification MLP is removed in ``w/o GauRec'' model. The ``w/o OffsetLoss'' and ``w/o ScalingLoss'' models contain rectification MLP but discard the regularization loss $\mathcal{L}_{offset}$ and scaling loss $\mathcal{L}_{scaling}$, respectively.

The ablation results are shown in Table~\ref{tab:ablation} and Figure~\ref{fig:ablation}. The results in Table~\ref{tab:ablation} show that the GauRec module can significantly improve PSNR and SSIM by adjusting Gaussian properties to learn more accurate non-rigid distortions and complex dynamic textures (such as clothes and hair). The result in the second column of Figure~\ref{fig:ablation} shows that removing the GauRec module results in burr textures on loose clothing. However, large offsets predicted by the GauRec module may affect the consistency of the dynamic avatar representation. To address this issue, we introduce the offset regularization loss. The comparison between the ``w/o OffsetLoss'' model and our full model in Figure~\ref{fig:ablation} shows that with the offset loss, our method can generate subtle Gaussian offsets to maintain the consistency of Gaussian splats and more accurately capture deformation details, resulting in improved rendering quality and color fidelity.

The scaling loss is to regularize the scaling of Gaussians and reduce Gaussians with elongated shapes. As shown in Table~\ref{tab:ablation}, without scaling loss, the PSNR on PeopleSnapshot decreases. The fourth column in Figure~\ref{fig:ablation} shows that elongated Gaussian splats cause severe rendering artifacts.

\begin{table}[h]
 \small
 \centering
  \caption{\textbf{Ablation Study on PeopleSnapshot~\cite{alldieck2018video}.} 
 The Gaussian rectification module and regularization losses improve reconstruction quality of clothed avatars. The best results are in bold.}
 \renewcommand{\arraystretch}{1.0}
 \begin{tabular}{@{}lccc}
 \toprule
 Metric:          
 & PSNR$\uparrow$
 & SSIM$\uparrow$
 & LPIPS$\downarrow$\\ \hline

  w/o GauRec
 & 31.92
 & 0.974
 & 0.017
\\
  w/o OffsetLoss
 & 32.48
 & \textbf{0.983}
 & 0.017
\\
  w/o ScalingLoss
 & 32.45
 & \textbf{0.983}
 & 0.017
\\
 Full model               
 & \textbf{32.51}
 & \textbf{0.983}
 & 0.017
\\
 \bottomrule
 \end{tabular}
 \label{tab:ablation}
\end{table}

\section{Conclusion and Discussion}

In this paper, we have proposed a hybrid representation for human avatar from monocular video based on mesh-embedded Gaussian splats. We utilize the explicit representation of the mesh to ensure correct shape and motion of avatar, and the implicit Gaussian splats located on mesh to render photorealistic appearance. To compensate for the limitation of linear transformations of LBS, we design a non-rigid deformation module that takes pose and Gaussian positions as input to further optimize the Gaussian properties, thereby learning the dynamic effects of flexible materials such as clothing and hair. Compared with the SoTA methods, our method achieves the best performance in both rendering quality and measurement indicators, and the reconstruction based on monocular video can further promote the application of avatars in multiple fields such as tele-presentation and virtual reality.

Our method learns non-rigid deformation based on mesh-guided Gaussian motion, and reduces reconstruction errors caused by inaccurate poses. However, due to the lack of meshes for clothing and facial expressions, our method has limited ability to model these details. In the future, we consider combining layered meshes and Gaussians to further improve the reconstruction of complex details such as the head, clothing, and hands.


\bibliographystyle{cas-model2-names}

\bibliography{cas-refs}



\end{document}